\documentclass[12pt]{article}

\usepackage{sbc-template}
\usepackage{graphicx,url}
\usepackage[utf8]{inputenc}
\usepackage[brazil]{babel}

\title{A RAG-Based Institutional Assistant}

\author{
  Gustavo Kuratomi$^{1}$\and
  Paulo Pirozelli$^{2}$\and
  Fabio G. Cozman$^{1}$\and
  Sarajane M. Peres$^{3}$
}

\address{
  Escola Politécnica\\
    $^{2}$Instituto de Estudos Avançados\\
  $^{3}$Escola de Artes, Ciências e Humanidades\\
  Universidade de São Paulo\\
  \email{\{gustavokpc,paulo.pirozelli.silva,fgcozman,sarajane\}@usp.br}
}

\begin{document} 

\maketitle

\begin{abstract}
Although large language models (LLMs) demonstrate strong text generation capabilities, they struggle in scenarios requiring access to structured knowledge bases or specific documents, limiting their effectiveness in knowledge-intensive tasks. To address this limitation, retrieval-augmented generation (RAG) models have been developed, enabling generative models to incorporate relevant document fragments into their inputs. In this paper, we design and evaluate a RAG-based virtual assistant specifically tailored for the University of São Paulo. Our system architecture comprises two key modules: a retriever and a generative model. We experiment with different types of models for both components, adjusting hyperparameters such as chunk size and the number of retrieved documents. Our optimal retriever model achieves a Top-5 accuracy of 30\%, while our most effective generative model scores 22.04\% against ground truth answers. Notably, when the correct document chunks are supplied to the LLMs, accuracy significantly improves to 54.02\%, an increase of over 30 percentage points. Conversely, without contextual input, performance declines to 13.68\%. These findings highlight the critical role of database access in enhancing LLM performance. They also reveal the limitations of current semantic search methods in accurately identifying relevant documents and underscore the ongoing challenges LLMs face in generating precise responses.
\end{abstract}

\section{Introduction}
Question answering (QA) has long been a central concern in natural language processing. Traditionally, models have been fine-tuned on question-answering datasets—such as SQuAD \cite{squad} and MS MARCO \cite{MSMarco}—following an initial phase of language model pretraining. With the advent of large language models (LLMs), standalone fine-tuning has mostly become obsolete, due to to these models' impressive in-context learning abilities \cite{bubeck2023sparksartificialgeneralintelligence}. LLMs generally perform satisfactorily on question-answering tasks, especially for factual questions, since they have likely encountered relevant information during training. However, LLMs still face a significant limitation: they do not have access to external knowledge bases, making it difficult to ensure they respond according to predefined answers. This limitation is particularly pressing when answers depend on documents that are private, frequently updated, or difficult to access.

Retrieval-Augmented Generation (RAG) \cite{RAG2020} is an architecture designed to address this problem. It consists of a modular structure where—in its standard version—two components are coupled together: a retriever module, responsible for finding relevant chunks of text, and a language generation module, which takes a question and the retrieved information to create an answer. This complex architecture allows better control over the information the model uses to answer questions, enables tracing back the source of answers, and disentangles the different steps in answer generation (retrieval and generation). More importantly, new documents can be constantly added to the database without the need for additional retraining.

In this paper, we present and evaluate a RAG-based application tailored for an educational context: a virtual assistant for the University of São Paulo (USP).\footnote{\url{https://www5.usp.br/}.} USP, one of the largest universities in Brazil, with nearly 100K students and a complex structure that includes multiple campuses, institutes, and courses, presents unique challenges in accessing information related to its institutional and regulatory aspects. This environment provides a valuable opportunity to build and assess a RAG system that can assist students, faculty, and staff in navigating USP’s extensive bureaucratic framework. Furthermore, it serves as an ideal scenario to compare the effectiveness of RAG-based systems with pure LLM approaches.

Our main contributions are as follows:

\begin{itemize}
    \item Developing a RAG-based QA system specifically designed to address the normative and institutional aspects of the University of São Paulo.
    \item Curating a comprehensive database of USP documents and constructing a QA dataset derived from these materials.
    \item Analyzing the impact of different system components (e.g., model type, chunk size) on answer quality.
\end{itemize}

All data and code are available on our GitHub page.\footnote{\url{https://github.com/gustavokpc/RetrievalAugmentedGeneration}.}

\section{Related Works}
Until recently, question answering was primarily addressed as a fine-tuning task, where a pretrained language model was adapted to a specific domain dataset, as exemplified by \cite{squad2.0}. However, with the rapid evolution of LLMs like GPT-4 \cite{openai2024gpt4}, Llama 2 \cite{touvron2023llama}, and Gemini \cite{geminiteam2024gemini}, this method has been largely replaced by prompt engineering. However, despite  their impressive ability to encode vast amounts of information, LLMs remain prone to hallucinations \cite{rawte2023survey}. This vulnerability raises concerns about their reliability, particularly in contexts that require high accuracy or involve sensitive information.

Retrieval-augmented generation (RAG) has emerged as an alternative to strictly parametric models \cite{RAG2020}. In this architecture, a retrieval model first gathers relevant information, which is then fed into a generative model to produce the final answer. This approach has the advantage of utilizing a knowledge base for information, allowing the generative model to concentrate on synthesizing the retrieved information into a coherent response. In the original RAG framework, both components were neural networks, and the system was trained end-to-end on a question-answering task. While the results were impressive, the authors noted that the learning process was fragile; the generator module could easily learn to ignore the retrieved texts, especially when dealing with initially random chunks.

A notable variant of RAG system, where components are built independently, has gained popularity in recent years. This modular approach facilitates the combination of various retrieval and generation modules. Typically, the embedding network employs a smaller, pre-trained model, while the generator utilizes an LLM. Over time, several enhancements to the RAG architecture have been proposed \cite{gao2024retrievalaugmented}, including preprocessing techniques such as query rewriting \cite{ma2023query}, the incorporation of metadata \cite{ilin2023advanced}, and the integration of diverse information sources like knowledge bases \cite{wang2023knowledgpt} and tabular data \cite{chen-etal-2020-hybridqa}. More advanced techniques include passage reranking to refine the retrieval process \cite{zhuang2023opensource} and iterative retrieval to address multihop problems \cite{shao2023enhancing}. In this paper, we employ this modular variant of the RAG architecture, testing different components for both the retrieval and generative modules.

\section{Dataset}
The University of São Paulo (USP) is one of the largest and most prestigious higher education institutions in Brazil. With an annual budget of R\$8.6 billion for 2024, it comprises 42 teaching and research units across 8 campuses in 9 cities. The main campus, known as the Armando de Salles Oliveira University City in São Paulo, covers nearly 3.7 million square meters—slightly larger than Central Park. USP offers 246 undergraduate and 229 graduate programs spanning all fields of knowledge, along with community-focused extension courses. Its personnel include 5K professors, 12K staff members, and 97K students enrolled in undergraduate and graduate programs. The university's extensive infrastructure features a radio station, research institutions like the Butantã Institute, museums such as the Museum of Ipiranga and the Museum of Modern Art, and hospitals, including the Hospital das Clínicas, the largest hospital complex in Latin America.\footnote{\url{https://uspdigital.usp.br/anuario/AnuarioControle}.}

This complex structure required a careful selection of documents. We focused exclusively on normative documents at the university level, excluding regulations specific to individual institutes, campuses, or departments. This choice was guided by the online availability of these documents and their broader relevance to our RAG system. We collected 866 documents from USP's website, covering the period from January 2023 to May 2024. These documents include a variety of norms, such as Historical Norms, the Statute, General Regulations, Resolutions, Ordinances, Regulations of the Bodies, and other Normatives. Together, they establish the foundational guidelines of USP, addressing topics like faculty organization and the university charter.

\begin{table}[h]
\centering
\caption{Statistics for the various databases. Chunk size is given in number of characters. Min., max., and avg. represent the minimum, maximum, and average number of words in each database.}
\label{tab:chunk_statistics}
\begin{tabular}{c|c|c|c|c}
\hline
\textbf{Chunk Size} & \textbf{Documents} & \textbf{Min. Size} & \textbf{Max. Size} & \textbf{Avg. Size} \\ 
\hline
2K & 4780 & 4 & 372 & 235.67  \\
4K & 2661 & 5 & 735 & 428.78  \\
8K & 1635 & 7 & 1401 & 697.25 \\
\hline
\end{tabular}
\end{table}

The documents were then divided into chunks to serve as input for the generative models. We experimented with three different chunk sizes: 2K, 4K, and 8K characters. Table \ref{tab:chunk_statistics} summarizes the statistics for these datasets. A common issue with chunking is that a relevant span of text may be split across chunks. To mitigate this, we incorporated overlapping characters,  including small consecutive portions of both the preceding and following chunks. We used a 1:10 ratio for the overlaps; for instance, with a 2K chunk size, we added 200 characters from both the preceding and subsequent chunks.

\begin{figure}[t]
 \centering
 \includegraphics[width=1\textwidth]{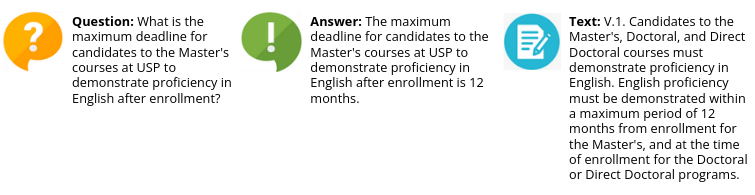}
 \caption{Example from the QA dataset, with question, answer, and context.}
 \label{fig:qa_rag}
\end{figure}

To evaluate this system, we developed a QA dataset based on the documents.  Questions were automatically generated using GPT-4 \cite{openai2024gpt4}, accessed through OpenAI's API. We started by randomly selecting chunks from the 2K-character database, as initial experiments indicated that this chunk size provided sufficient context for generating high-quality, realistic questions. We then prompted GPT-4 to generate a question based on the input text. The prompt included a brief overview of the University of São Paulo and specific instructions, such as, ``create self-contained questions that can be understood without direct reference to the text''. The model returned a string containing three elements: the question, the answer, and the text span where the answer could be found. We separated these elements using a regex strategy. Finally, all questions were manually reviewed to ensure that no two chunks in the database could answer a single question. Although this setup is somewhat artificial, limiting answers to a single chunk simplifies the evaluation of the retrieval system. The resulting dataset consists of 592 questions. Figure \ref{fig:qa_rag} shows an example question from the dataset.

Since the documents served as the basis for question generation, the questions often shared similar vocabulary with their sources. This created an unrealistic scenario, where identifying relevant texts and extracting answers became relatively easy due to the overlapping terms. To address this, we developed a second version of our QA dataset using GPT-4. In this version, we prompted the model to paraphrase the original questions. The input included the question, answer, and supporting text, with instructions to reformulate the question in a manner as different as possible while preserving the relevant content.

\section{Architecture}
\begin{figure}[t]
 \centering
 \includegraphics[width=0.8\textwidth]{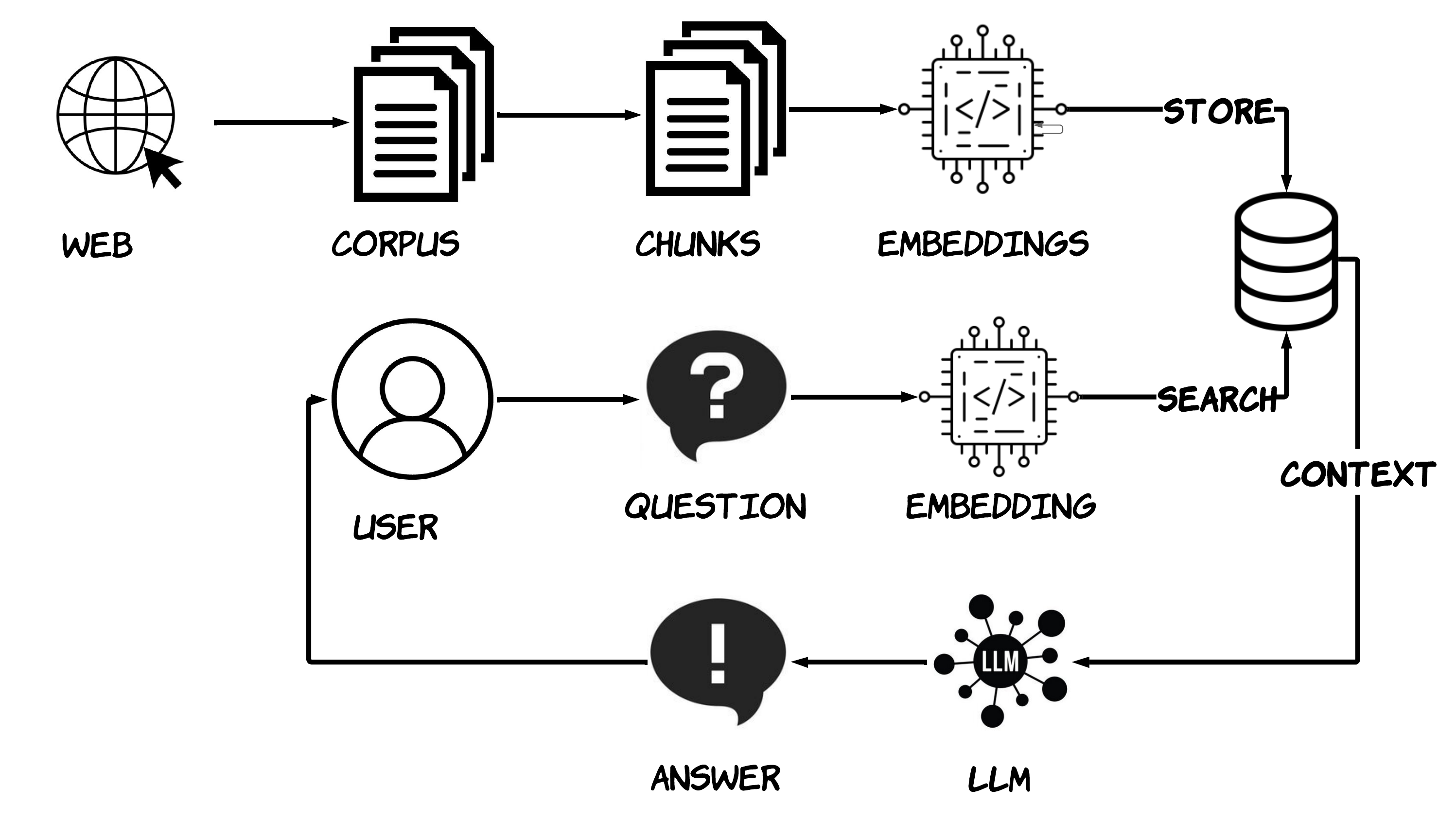}
 \caption{Overview of the system's architecture.}
 \label{fig:rag_overview}
\end{figure}

To build our assistant, we opted for a RAG system, as shown in Figure \ref{fig:rag_overview}. This choice was driven by two main factors. First, the knowledge base comprises a dynamic corpus of documents, with new entries added regularly, which would require constant retraining if using an end-to-end model. Second, the questions may address sensitive topics, such as eligibility for financial aid, where precision in both information retrieval and response generation is crucial. A modular architecture is better suited for such scenarios, with one component dedicated to retrieving relevant documents and another focused on generating accurate responses. This separation of tasks provides greater control and accuracy at each stage of the process.

The system is composed of two main modules: a retriever and a generative model. The retriever ranks text chunks based on their similarity to the user's question. In a one-time process, the text chunks are first converted into vector representations, or embeddings, which are then stored in a static vector database. When a query is made, the system generates an embedding for the question using the same model. The similarity between the query and each chunk is then calculated by taking the dot product of their vectors, following a method similar to DPR \cite{karpukhin2020dense}.\footnote{We also experimented with cosine similarity but observed no significant difference.} The model subsequently returns a ranked list of chunks based on their similarity scores.

The generative model is tasked with producing the final answer from the retrieved chunks. To do this, we construct a prompt that includes the user's question and the \textit{k} most similar chunks, presented in natural language. The prompt also provides relevant information about the University of São Paulo, along with specific instructions for the task. Using this input, the model generates a response to the question. To further enrich the context, metadata such as the document title and date is included with each chunk.

The coordination of all these components was managed using LangChain,\footnote{\url{https://www.langchain.com/}.} a widely-used framework for building applications powered by LLMs. In our system, LangChain was responsible for: i) loading the LLMs, ii) querying the database, and iii) conveying information between the modules. For storing embeddings, we employed FAISS, a library developed by Meta for efficient similarity search.\footnote{\url{https://faiss.ai/}.}

\section{Experiments}
Based on the modular aspect of our system, evaluations were conducted separately for each module.

\subsection{Retrieval}
Working with documents in Portuguese posed a challenge due to the limited availability of embedding models. To address this, we opted for multilingual models that support Portuguese. These text embedding models are all based on Sentence-BERT \cite{reimers2019sentencebert}, which employs siamese BERT networks to generate semantic representations of documents.\footnote{\url{https://www.sbert.net/docs/sentence\_transformer/pretrained\_models.html}.} Four models were used: paraphrase-multilingual-MiniLM-L12-v2, paraphrase-multilingual-mpnet-base-v2, distiluse-base-multilingual-cased-v1, and distiluse-base-multilingual-cased-v2. Additionally, we tested BM25 \cite{robertson2009probabilistic}, a keyword-based retriever, as a baseline. BM25 is a ranking function estimates document relevance to a search query based on word occurrences. It improves upon the traditional TF-IDF (term frequency-inverse document frequency) approach by incorporating document length normalization, which favors shorter documents.  Lastly, we included a random baseline, calculated as $\frac{k}{C}$, where $k$ is the number of chunks selected as relevant, and $C$ is the total number of chunks in the database.

\begin{table}[h!]
\centering
\caption{Top-1 and Top-5 accuracy for different models in the retrieval task. The top rows display the accuracy for the original questions, while the bottom rows present the accuracy for the paraphrased questions. The best results for each case are highlighted in bold.}
\label{tab:retrieval_original}
\begin{tabular}{l|ll|ll|ll}
\hline
 & \multicolumn{2}{c|}{\textbf{2K}} & \multicolumn{2}{c|}{\textbf{4K}} & \multicolumn{2}{c}{\textbf{8K}} \\ 
\hline
\textbf{Model} & \textbf{Top-1} & \textbf{Top-5} & \textbf{Top-1} & \textbf{Top-5 }& \textbf{Top-1} & \textbf{Top-5} \\ 
\hline
MiniLM-L12-v2 & 0.23 & 0.40 & 0.10 & 0.19 & 0.05 & 0.11 \\
Mpnet-base-v2 & 0.23 & 0.36 & 0.10 & 0.19 & 0.07 & 0.12 \\
Distiluse-v1 & 0.18 & 0.36 & 0.11 & 0.22 & 0.08 & 0.16 \\
Distiluse-v2 & 0.12 & 0.21 & 0.07 & 0.14 & 0.05 & 0.10 \\
BM25 & \textbf{0.31 } & \textbf{0.51} & \textbf{0.31} & \textbf{0.54} & \textbf{0.35} & \textbf{0.57} \\
Random & 0.00 & 0.00 & 0.00 & 0.00 & 0.00 & 0.00 \\ 
\hline
\hline
MiniLM-L12-v2 & 0.13 & 0.27 & 0.06 & 0.13 & 0.05 & 0.11 \\
Mpnet-base-v2 & \textbf{0.16} & \textbf{0.30} & 0.07 & 0.14 & 0.06 & 0.10 \\
Distiluse-v1 & 0.13 & 0.27 & 0.08 & 0.17 & 0.07 & 0.13 \\
Distiluse-v2 & 0.08 & 0.15 & 0.05 & 0.11 & 0.03 & 0.08 \\
BM25 & 0.15 & \textbf{0.30} & \textbf{0.17} & \textbf{0.32} & \textbf{0.19} & \textbf{0.37} \\
Random & 0.00 & 0.00 & 0.00 & 0.00 & 0.00 & 0.00 \\ 
\hline
\end{tabular}
\end{table}

\begin{figure}[ht]
    \centering
    \includegraphics[width=1\textwidth]{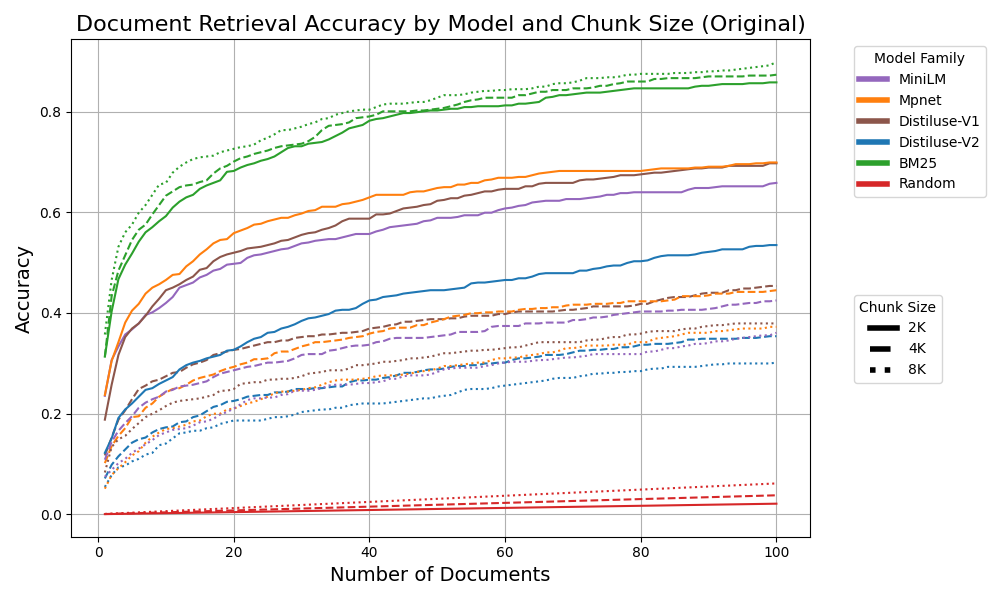}
    \caption{Top-k accuracy for retriever models using the original questions. Different models are represented by color, and chunk sizes are distinguished by line type.}
    \label{fig:retrieval_original}
\end{figure}

To identify the best retriever, we varied two hyperparameters: the model type and chunk size. Our primary evaluation metric was top-\textit{k}, which measures the percentage of cases where the correct chunk is within the top \textit{k} ranked results. The top rows in Table \ref{tab:retrieval_original} present the Top-1 and Top-5 values using the original questions. Figure \ref{fig:retrieval_original} illustrates the performance of different models across 1 to 100 retrieved documents, according to chunk size. BM25 significantly outperformed the neural models, showing consistent results across all chunk sizes. For the neural models, smaller chunk sizes were easier to retrieve, possibly because larger documents were more challenging to encode effectively. The random retriever's performance was approximately zero.

\begin{figure}[ht]
    \centering
    \includegraphics[width=1\textwidth]{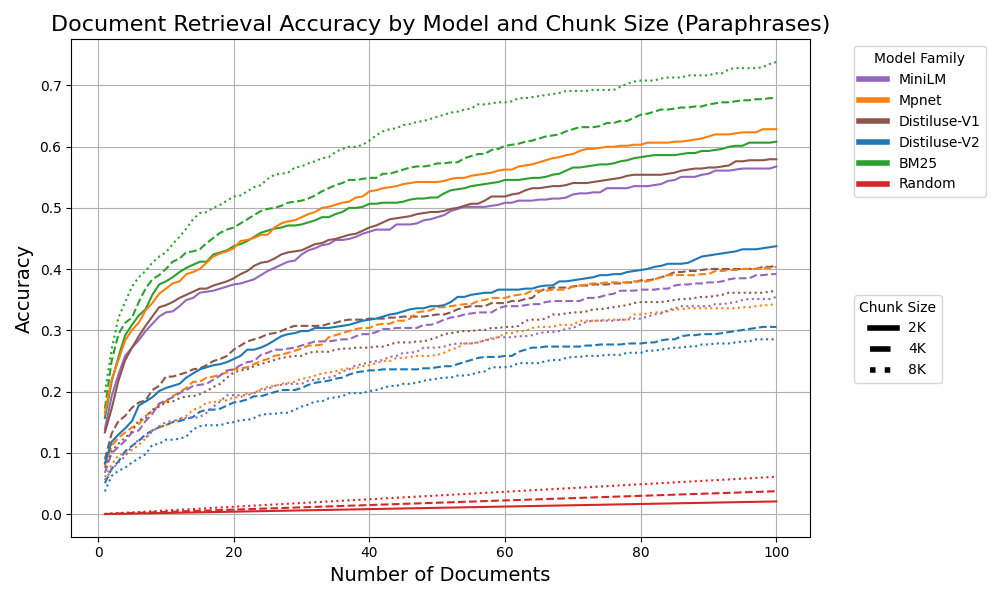}
    \caption{Top-k accuracy for retriever models using the paraphrased questions. Different models are represented by color, and chunk sizes are distinguished by line type.}
    \label{fig:retrieval_paraphrases}
\end{figure}

BM25's success can be partly attributed to the fact that the questions in our dataset were generated from the same texts used in the retrieval process. As a result, the questions naturally shared vocabulary with the supporting documents, giving BM25 an inherent advantage. More importantly, this setup fails to capture the linguistic variability present in real-world applications, where users' queries often differ in phrasing from the reference documents. To provide a more balanced evaluation, we tested the same models on our dataset of paraphrased questions. The bottom rows of Table \ref{tab:retrieval_original} present the Top-1 and Top-5 accuracy for the different models and chunk sizes using the paraphrased questions. Performance dropped in all cases, but the decrease was less pronounced for embedding models. Interestingly, for the smallest chunk size, Mpnet slightly outperformed BM25, suggesting that while semantic search may be limited in environments with minimal lexical and syntactic variation, it becomes advantageous in more realistic scenarios. Figure \ref{fig:retrieval_paraphrases} presents the results for the different models across 1 to 100 retrieved documents, separated by chunk size.


\subsection{Generation}\label{sec:generation}
The second experiment aimed to assess the quality of answers produced by different LLMs. For each question, we retrieved the most relevant text chunks and provided them to the LLMs, along with the question and specific instructions.

In all cases, we used the same retriever, Mpnet, with a chunk size of 2K tokens, as this model achieved the highest score in the 2K database. This choice was driven by several factors. First, there is an inherent trade-off between the number and size of chunks: larger chunk sizes reduce the number of chunks that can be provided to the model. Second, all retriever models performed best on the 2K database in terms of top-\textit{k} accuracy. Additionally, evidence suggests that LLMs are not robust to long input contexts \cite{liu2024lost, li2024long}.

In this experiment, we varied two hyperparameters: the model type and the number of chunks passed to the model. We evaluated both open-source and proprietary LLMs, including GPT-3.5-turbo (ChatGPT) \cite{openai2024gpt4}, Llama-3-8B-Instruct,\footnote{\url{https://ai.meta.com/blog/meta-llama-3/}.} Mixtral-8x7B-Instruct \cite{jiang2023mistral7b}, and Sabia-2-medium \cite{maritaca2024sabia2}. At this stage, we opted to use GPT-3.5, reserving GPT-4 for dataset creation and the qualitative evaluation described later. 
For the number of chunks, we experimented with $k = [3, 5, 8]$.

Evaluating generative text is inherently challenging, so we utilized multiple metrics to provide a comprehensive assessment of the system's performance.  The first metric was the F1-score, which combines precision and recall to measure the accuracy of the generated answers. In this context, precision and recall were calculated based on the number of shared words between the predicted and reference answers, focusing on word similarity. The second metric was the cosine similarity between the embeddings of the predicted and reference answers, using Mpnet as the embedding model. This approach allowed us to assess semantic similarity, even when the answers were phrased differently. Lastly, we conducted a qualitative evaluation by prompting GPT-4 to rate the generated answers on a predefined scale: \texttt{Totally correct} for answers that completly match the correct response and appropriately address the question based on the supporting text; \texttt{Mostly correct} for answers that are largely accurate but contain minor errors or omissions; and \texttt{Incorrect} for answers that are wrong or fail to adequately address the question. All metrics were then rescaled to a 0-100 range.

The results for the various models are summarized in Table \ref{tab:generative_models}.  The top rows display the performance metrics for all generated answers, irrespective of whether the correct chunk was included among the prompts. Among the evaluated models, GPT-3.5 demonstrated the best overall performance across all metrics. Additionally, a slight improvement was observed when more data chunks were provided, suggesting that increased context can enhance the accuracy of the generated outputs. 

\begin{table}[h]
\centering
\caption{F1-score, cosine similarity, and LLM scores for various language models across different numbers of retrieved documents ($k$). The top rows display performance metrics for all responses, including those without the correct chunk in the prompt. The middle rows focus on cases where the correct chunk was included in the prompt. The bottom rows show performance when no context is provided ($k=0$). The best results are highlighted in bold.}
\label{tab:generative_models}
\begin{tabular}{l|lll|lll|lll}
\hline
 & \multicolumn{3}{c|}{\textbf{k=3}} & \multicolumn{3}{c|}{\textbf{k=5}} & \multicolumn{3}{c}{\textbf{k=8}} \\ 
\hline
\textbf{Model} & \textbf{F1} & \textbf{Cos.} & \textbf{LLM} & \textbf{F1} & \textbf{Cos.} & \textbf{LLM} & \textbf{F1} & \textbf{Cos.} & \textbf{LLM} \\ 
\hline
GPT-3.5 & \textbf{34.71}  & \textbf{89.18}  & \textbf{19.25 } &  \textbf{36.31} &  \textbf{89.43} & \textbf{21.03}  &  \textbf{36.24} & \textbf{89.56}  & \textbf{22.04}  \\
Llama-3 & 17.96  &  85.44 & 15.03  &  19.05 & 87.61  & 16.55  & 19.85  & 88.10  & 17.56  \\
Mixtral &  29.85 & 88.35 & 15.54 &  30.97 & 88.48  & 16.21  & 32.83   & 88.96  &  20.43 \\
Sabia-2 &  28.48 &  88.18  & 15.03  & 29.76  & 88.44  &  18.91 & 30.93  & 88.49  & 21.11  \\
\hline
\hline 
GPT-3.5 & \textbf{47.50}   & \textbf{92.99} & \textbf{54.02} & \textbf{48.87}  & \textbf{92.93}  & \textbf{51.96}  & \textbf{47.08}  & \textbf{ 92.61} &  \textbf{50.24} \\
Llama-3 & 23.61  & 89.94 & 46.30 & 24.61  & 90.66  &  42.97 &  24.18 &  90.96 & 40.19  \\
Mixtral & 41.76  & 92.25 & 45.97 &  41.82  & 91.82 &  43.53 & 42.28  &  91.85 &  48.28 \\
Sabia-2 &  38.30 & 91.23 & 44.96 & 39.19 & 91.10 & 47.19  &  39.11 & 91.07  &  47.54 \\
\hline
\hline
GPT-3.5 & & &  & \textbf{35.08} & \textbf{88.90} & \textbf{13.68} & & & \\
Llama-3 & & &  &  15.36 & 87.32 & 3.04 \\
Mixtral & & &  &  25.41 & 87.20 & 4.89\\
Sabia-2 & & &  & 27.02 & 87.36 & 6.58 \\
\hline
\end{tabular}
\end{table}

The results change considerably when we focus solely on questions where the correct text is included among the provided chunks (194, 178, and 204 answers for $k = 3, 5$, and $8$, respectively). The middle block of rows in Table \ref{tab:generative_models} presents the performance metrics for these specific cases. All metrics show substantial improvement, most notably in the LLM evaluation. On average, the answers improved by approximately 30 points according to this metric and around 10 points for the F1-score. However, the cosine similarity remained relatively unchanged, indicating that it may not be an ideal metric for this type of generative evaluation. As expected, when the correct chunk is included in the prompt, providing more irrelevant chunks tends to decrease the model's performance. Despite these gains, even our best model, GPT-3.5, still exhibits significant room for improvement.

\begin{figure}[h]
 \centering
 \includegraphics[width=0.6\textwidth]{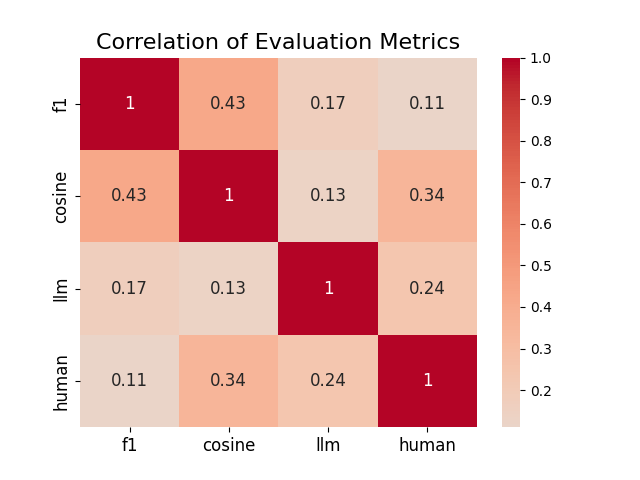}
 \caption{Correlation matrix showing the relationships between different evaluation metrics (F1, cosine similarity, LLM, and human assessment).}
 \label{fig:correlation}
\end{figure}

To assess the adequacy of the metrics, we performed a manual evaluation of the answers generated by Llama-3 in the $k=5$ database. Only questions where the prompt included the correct chunk were considered. Using the same criteria as those applied to the GPT-4 model, the manual evaluation resulted in a score of 88.20\%. Despite this high score, we observed several issues in the responses. The model occasionally mixed Portuguese and English, leading to inconsistencies in language use. For questions containing multiple queries, it sometimes failed to address all parts. In a few instances, it provided random responses, completely ignoring the context texts. In two cases, the model did not produce any response. Figure \ref{fig:correlation} illustrates the correlation between the different evaluation metrics.

Finally, we evaluated the performance of the LLMs without retrieving any chunks from an external database.  This comparison aimed to quantify the benefits of using a RAG architecture versus relying solely on direct prompting of an LLM. The results of these evaluations are shown in the bottom rows of Tabl \ref{tab:generative_models}.\footnote{Only one set of results is provided, as there is no variation in context size (i.e., $k=0$).} The findings clearly demonstrate that all models perform better when they have access to a structured database. According to the GPT-4 evaluations, model performance drops by 7.35 (GPT-3.5) to 13.51 points (Llama3) when no context is provided, considering all questions; and by approximately 40 points when focusing only on questions with correctly retrieved chunks. These results highlight the significant contribution of the RAG architecture to the system's overall effectiveness.

\section{Interface}
To facilitate the use of our RAG-based virtual assistant, we developed a web application using Streamlit, an open-source Python framework for building interactive web applications.\footnote{\url{https://streamlit.io/}} Additionally, to enhance the user experience, we implemented a select box feature that allows users to switch between different ``Chat Sessions'', each maintaining its own history of messages and interactions. For message storage, we utilized Python’s \texttt{shelve} library, which provides a dictionary-like object where keys are strings and values are Python objects, all stored locally on the system. 


\section{Conclusion}
We developed and analyzed a virtual assistant designed to answer questions about the University of São Paulo, using a modular, RAG-based architecture that combines a retriever and a generative model. Alongside compiling a corpus of official documents and creating a QA dataset, we conducted extensive experiments with various retrieval and generative models, exploring different hyperparameters such as chunk size and the number of chunks provided. The results indicate that our system's performance is significantly hampered by retrieval failures, primarily due to the limitations of multilingual embedding models, as well as inherent challenges within current LLMs.


\section{Limitations}
While our RAG-based system for question answering has demonstrated promising results, several enhancements could further improve its robustness and reliability. The most immediate step would be to expand our database by incorporating documents from a wider range of sources, including various institutes and departments within the University of São Paulo.  For the QA dataset, integrating questions posed by actual users and including multi-hop questions—those that require linking information across different text segments—would greatly enhance the system's ability to handle more complex queries. Another key area for improvement involves systematic evaluations by experts, particularly academic staff, whose feedback would provide a more rigorous assessment of the model’s accuracy and reliability. Finally, testing the system with real users—including students, faculty, and non-affiliated individuals—will be essential for evaluating its practical utility and overall effectiveness.

\section*{Acknowledgement}
This work was carried out at the Center for Artificial Intelligence (C4AI-USP),  with support by the S\~ao Paulo Research Foundation (FAPESP grant 2019/07665-4)  and by the IBM Corporation. Paulo is supported by the FAPESP grant \#2019/26762-0. Fabio is partially supported by CNPq \#305753/2022-3.

\bibliographystyle{sbc}
\bibliography{sbc-template}

\end{document}